\pdfoutput=1

\documentclass[11pt]{article}

\usepackage[preprint]{acl}
\usepackage{times}
\usepackage{latexsym}
\usepackage{amsmath}
\usepackage{amsfonts}
\usepackage{pgfplots}
\pgfplotsset{compat=1.18}
\usepackage{subcaption}
\usepackage{booktabs}

\usepackage{enumerate}

\usepackage[T1]{fontenc}

\usepackage[utf8]{inputenc}

\usepackage{microtype}

\usepackage{inconsolata}

\usepackage{graphicx}

\title{BcQLM: Efficient Vision-Language Understanding with Distilled Q-Gated Cross-Modal Fusion}

\author{Sike Xiang \and Shuang Chen \and Amir Atapour-Abarghouei\thanks{Corresponding author.}\\
Department of Computer Science, Durham University\\
\texttt{\{sike.xiang, shuang.chen, amir.atapour-abarghouei\}@durham.ac.uk}
}

\begin{document}
\maketitle
\begin{abstract}
As multimodal large language models (MLLMs) advance, their large-scale architectures pose challenges for deployment in resource-constrained environments. In the age of large models, where energy efficiency, computational scalability and environmental sustainability are paramount, the development of lightweight and high-performance models is critical for real-world applications. As such, we propose a lightweight MLLM framework for end-to-end visual question answering. Our proposed approach centres on BreezeCLIP, a compact yet powerful vision-language encoder optimised for efficient multimodal understanding. 
With only 1.2 billion parameters overall, our model significantly reduces computational cost while achieving performance comparable to standard-size MLLMs. Experiments conducted on multiple datasets further validate its effectiveness in balancing accuracy and efficiency. The modular and extensible design enables generalisation to broader multimodal tasks. The proposed lightweight vision-language framework is denoted as BcQLM (BreezeCLIP-enhanced Q-Gated Multimodal Language Model). It offers a promising path toward deployable MLLMs under practical hardware constraints. The source code is available at \url{https://github.com/thico0224/BcQLM}.

\end{abstract}

\section{Introduction}

Multimodal learning integrates visual and linguistic modalities, which has become a key direction in building more general and human-like AI systems. By combining information from different modalities, such systems can better understand complex real-world scenarios and perform more advanced reasoning and decision-making tasks.

Among various multimodal tasks, Visual Question Answering (VQA) has emerged as a widely adopted and representative benchmark. VQA requires models to comprehend an image, understand a corresponding natural language question and generate an appropriate answer \cite{VQA}. It poses significant challenges in cross-modal semantic alignment, contextual reasoning and fine-grained visual understanding \cite{butd, blip, VinVL}, and thus serves as an important indicator of multimodal intelligence.

To effectively address the inherent challenges of VQA and other complex multimodal tasks, mainstream approaches have predominantly adopted a two-stage architecture. Pretrained encoders such as CLIP \cite{clip}, ViT \cite{vit}, and BERT \cite{bert} are first used to extract semantic features from images and texts, providing strong unimodal representations and initial cross-modal alignment through large-scale pretraining. These features are then passed to large language models (LLM), which serve as decoders to perform reasoning and generate answers. This paradigm has been adopted by several representative systems, including BLIP-2, Qwen and LLaVA \cite{blip2, qwen, llava}. These models integrate visual understanding with language modelling, which enables them to generate contextually relevant and accurate answers across a wide range of question types.

However, existing multimodal large models often suffer from large parameter sizes and high computational costs, limiting their deployment on edge devices. This hinders practical use in scenarios requiring efficient on-device inference, such as healthcare \cite{ChatCAD}, remote education \cite{Education}, disaster response \cite{FloodLense} and assistive technologies \cite{PaLM-E}. In these settings, cloud-based solutions may be infeasible due to connectivity or latency issues \cite{10640100}, underscoring the need for lightweight yet capable multimodal systems.

To address this issue, we propose BcQLM (BreezeCLIP-enhanced Q-Gated Multimodal Language Model).  BreezeCLIP is a lightweight vision–language encoder designed to significantly reduce computational costs while enabling efficient multimodal representation. It replaces the original BERT and ViT backbones in CLIP with compact transformer modules inspired by the inverted bottleneck design \cite{MobileNetV2}. To train BreezeCLIP, we construct image–text pairs from the GQA dataset \cite{GQA}, leveraging its rich scene graph annotations that provide detailed object relationships and spatial context. We adopt a dual training strategy: contrastive learning ensures visual and textual features are well aligned in a shared embedding space, while knowledge distillation from a CLIP teacher \cite{clip} transfers high-level semantic alignment to BreezeCLIP. This enables BreezeCLIP to achieve strong performance despite its lightweight architecture.

A lightweight decoder based on LLaMA-3.2-1B \cite{llama3} is first employed to generate answers conditioned on fused multimodal features, achieving strong performance with significantly reduced computational overhead. Building on this, we introduce a Q-Gated Cross-Modal Fusion Module that dynamically adjusts the contribution of visual and textual features according to the input question, which enables fine-grained, question-aware interaction between modalities. These components together constitute an efficient end-to-end multimodal question answering system designed for real-world, resource-constrained deployment scenarios.

Our primary contributions are thus as follows:

\begin{enumerate}[(i)]   
\vspace{-0.1cm}
\item We propose BcQLM, involving a compact BreezeCLIP by distillation learning and a Q-Gated Cross-Modal Fusion Module. Comprehensive experiments demonstrate that our BreezeCLIP is able to effectively preserve vision-language alignment capabilities under a tiny model setting.
\vspace{-0.3cm}
\item We propose a Q-Gated Cross-Modal Fusion Module to enable fine-grained and adaptive multimodal fusion.
\vspace{-0.3cm}
\item Our BreezeCLIP only contains 1.2B parameters (10\% of SoTA method), which achieves performance comparable to several standard-sized MLLMs with much higher parameter counts on VQA tasks.

\end{enumerate}

\begin{figure*}[t]
  \centering
  \includegraphics[width=\textwidth]{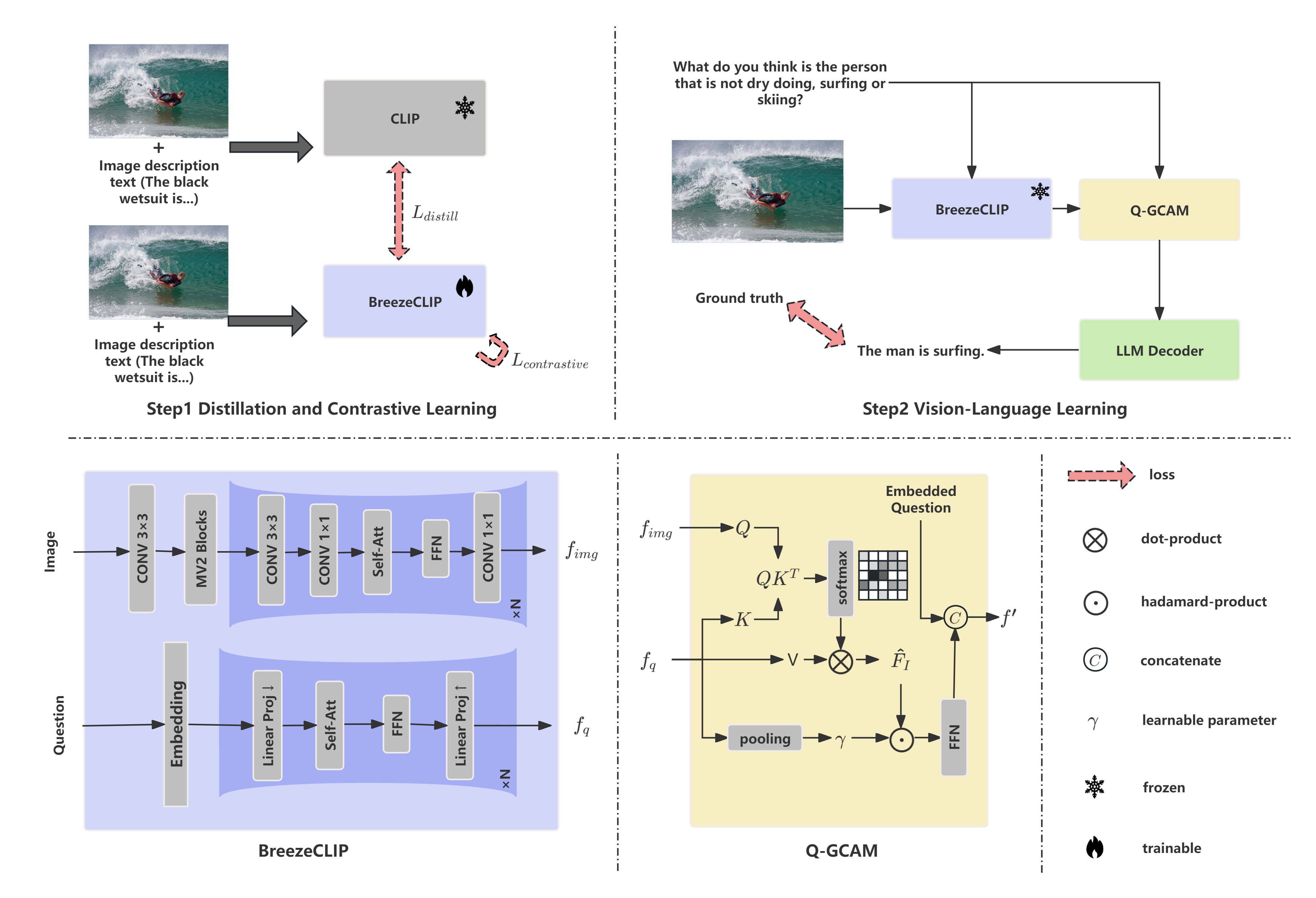}

  \caption{An overview of the proposed framework.
Training consists of two stages: (1) BreezeCLIP encoder is trained using contrastive ($\mathcal{L}{\text{contrastive}}$) and distillation loss ($\mathcal{L}{\text{distill}}$) from the CLIP teacher; (2) the Q-GCAM module fuses image and question features to generate answers via a lightweight LLM decoder, optimised with cross-entropy loss ($\mathcal{L}_{\text{CE}}$). The bottom diagrams illustrate BreezeCLIP and Q-GCAM architectures.}
  \label{1} 
\end{figure*}

\section{Related Work}
We consider prior work leading to recent advances in MLLMs, with connections to multimodal fusion techniques and vision-language tasks, such as VQA, highly relevant to our work.

\subsection{Visual Question Answering}

Visual Question Answering (VQA) is a prominent multimodal task that requires a model to answer natural language questions based on the content from an image. It requires both visual perception and language understanding, as well as reasoning across the two modalities. First introduced by Antol et al. \cite{VQA}, the VQA task has become a widely used benchmark for evaluating multimodal intelligence. Subsequent work by Goyal et al. \cite{vqav2} extended the original dataset with balanced answer distributions, mitigating language bias and enabling more robust evaluation.

Early VQA approaches extract CNN-based image features and LSTM/GRU question embeddings, fusing them via concatenation, bilinear pooling or co-attention (e.g., MCB \cite{mcb}); later attention-based models like BAN \cite{ban} and Bottom-Up Top-Down \cite{butd} enhance fine-grained reasoning and cross-modal alignment. These early methods achieved some success but relied on task-specific fusion and lacked generalisation. Recent work leverages pretrained large language models and generative architectures to enable broader reasoning.

\subsection{Vision-Language Pretraining \& LLM Generation}

Advances in multimodal representation learning focused on aligning visual and textual modalities through joint embedding spaces. One of the most representative models, CLIP \cite{clip}, learns such a space by contrastively training an image encoder and a text encoder on large-scale image-text pairs. Subsequent models such as ViLT \cite{vilt}, ALBEF \cite{albef}, and BLIP \cite{blip} further enhance cross-modal alignment by incorporating attention-based fusion mechanisms and multi-task pretraining objectives. These models demonstrate strong transferability to a wide range of downstream tasks, including VQA, image-text retrieval \cite{flair2024} and caption generation \cite{SynthVLM}.
Such dual-encoder and fusion-based models adopt architectures where visual and textual inputs are first encoded independently and then fused via attention modules or interaction layers. While they excel at semantic alignment, they remain limited in their ability to support complex reasoning or generate language that relies on common sense and broader world knowledge not captured by the training corpus. 

In response to these limitations, recent studies have introduced pretrained LLMs into multimodal systems, marking a shift toward generative architectures. Unlike traditional approaches that rely solely on vision-language encoders, models such as BLIP-2 \cite{blip2} and LLaVA \cite{llava} integrate visual features into frozen or lightly tuned LLMs, treating the language model as the central reasoning and generation component. 
Visual inputs are typically embedded as prompts or special tokens to guide multimodal interaction. These models demonstrate superior performance in open-ended and knowledge-intensive tasks like VQA and produce more coherent and contextually relevant answers.
This shift from traditional dual-encoder models to LLM-centred generative architectures brings improved reasoning and generation capabilities, but high computational costs and large model sizes remain major obstacles for practical deployment. Recent works such as AdapterCLIP \cite{adapterclip} and LoRA \cite{lora} mitigate this issue via lightweight adapters or low-rank tuning, but still rely on large fixed backbones and show limited flexibility in complex generation tasks.
Our work advances this direction by emphasising lightweight and efficient modelling through the introduction of BreezeCLIP, coupled with gated cross-modal fusion and a LLaMA-based decoder.

\section{Method}

The proposed BcQLM (BreezeCLIP-enhanced Q-Gated Multimodal Language Model) is a lightweight multimodal large language model designed for visual question answering under resource constraints, to support effective multimodal reasoning in resource-limited environments. The architecture comprises three main components. First, a compact vision-language encoder, BreezeCLIP, is proposed to extract semantic features from both images and text. BreezeCLIP is trained using contrastive learning and knowledge distillation from a CLIP teacher to ensure efficient representation learning. Next, we propose a question-guided cross-modal attention module (Q-GCAM) to dynamically fuse visual and textual features based on the semantics of the input question. Finally, a lightweight language decoder based on LLaMA-3.2-1B generates the final answer. An overview of the framework is shown in Figure~\ref{1}. It follows a two-stage design: (1) BreezeCLIP is first pretrained via distillation and contrastive learning; and (2) the pretrained BreezeCLIP is then used as a frozen feature extractor, working with Q-GCAM and an LLaMA-3.2-1B decoder to perform vision-language generation.
This balances performance and efficiency, and enables BcQLM to support effective multimodal reasoning in resource-limited environments.

\subsection{Breezeclip}

To enable efficient vision-language representation learning within our framework, we design BreezeCLIP, a lightweight dual-encoder model that significantly reduces computational costs while effectively aligning visual and textual modalities.

\textbf{Feature Encoder:}  
The text encoder in BreezeCLIP adopts a lightweight Transformer to enable efficient language modelling in resource-constrained settings, as shown in Figure~\ref{1}. The architecture adopts the same 24-layer Transformer structure as BERT-Large. To reduce the number of computation parameters while preserving model capacity, we incorporate a bottleneck structure \cite{bertbottleneck} into each layer: token embeddings are first projected to lower dimensions, where attention and feed-forward computations are performed and then restored via a linear projection. Residual connections and layer normalisation are applied to ensure training stability. Furthermore, we adopt a factorised embedding strategy to reduce the size of the vocabulary embedding matrix and apply layer-wise knowledge distillation to preserve semantic representation quality. The hidden states from the last Transformer layer are retained as token-level features without pooling, preserving fine-grained textual semantics for subsequent fusion.

The image encoder combines convolutional operations with lightweight Transformer modules, following the bottleneck principle \cite{vitbottleneck}. This structure enables efficient local feature extraction while maintaining the ability to capture global context with minimal computational overhead. As shown in Figure~\ref{1}, the input image is first passed through a convolutional stem and a set of lightweight MobileNetV2-style bottleneck blocks \cite{MobileNetV2} to extract early features. The resulting feature maps are then processed by three Transformer blocks operating at progressively smaller spatial resolutions (e.g., $14\times14$ and $7\times7$). Each block consists of a local convolutional module followed by a stack of Transformer encoder layers — specifically, 2, 4, and 3 layers in the three blocks, respectively, totalling 9 self-attention layers. 
In each Transformer block, the input feature map is processed by a $3\times3$ convolution to capture local context, followed by a $1\times1$ convolution to reduce channel dimensions. The resulting features are flattened into non-overlapping patches and passed through the Transformer encoder for global modelling. The output is then reshaped back into the 2D spatial layout and fused with the original representation using a $1\times1$ convolution. Finally, the output feature map is flattened across spatial dimensions into a patch-level embedding sequence to produce a three-dimensional representation aligned with the text token embeddings for downstream cross-modal interaction. 

The constructed BreezeCLIP achieves a total parameter reduction of approximately 80\% (from 151M to 31M), which significantly improves deployment efficiency in resource-constrained environments compared to standard CLIP.


\textbf{Distillation and Contrastive Learning.}
A joint optimisation scheme that unifies contrastive learning and knowledge distillation is adopted to strengthen alignment across modalities in a lightweight setting. This enables the student model to learn from both raw image-text pairs and the supervision of a large pretrained teacher model.

We construct image-text pairs by leveraging the scene graphs provided in the GQA dataset. Each image is associated with a structured graph that specifies object categories, spatial relationships (e.g., ``a book is on the right of the table''), and attributes (e.g., ``the book is red''), as shown in Appendix Figure~\ref{fig:appendix-sg}. We convert these graph structures into informative language descriptions to serve as textual counterparts to the original images.


Given image-text pairs, the inputs are encoded using the student model's image and text encoders, resulting in feature embeddings $I_s$ and $T_s$. A contrastive InfoNCE loss \cite{InfoNCE} is used to bring matched pairs closer and push mismatched pairs apart in the shared embedding space:

{\small
\begin{equation}
\mathcal{L}_{\mathrm{contrast}} = \frac{1}{\alpha} \left( \mathrm{CE}\left( \frac{I_s T_s^\top}{\tau}, \mathbf{y} \right) + \mathrm{CE}\left( \frac{T_s I_s^\top}{\tau}, \mathbf{y} \right) \right),
\label{eq:contrast}
\end{equation}
}
where $\tau$ is a temperature coefficient, $\mathbf{y}$ denotes the index labels of positive pairs and $\alpha$ balances the contributions of the two directional terms.

\begin{table*}[ht]
  \centering
  
  \begin{tabular*}{\textwidth}{@{\extracolsep{\fill}}
      llc    
      c||     
      ccc   
    }
    \toprule
    \textbf{Method} & \textbf{LLM} & \textbf{Param. }
      & \textbf{Res.}
      & \textbf{GQA} & \textbf{VQA$^V$$^2$} & \textbf{VisWiz} \\
    \midrule
    BLIP-2 (2023)        
      & Vicuna-13B  & 13.5  & 224  & 44.0   & 65.0   & 19.6    \\
    InstructBLIP (2023) 
      & Vicuna-7B   & 7.5  & 224  & 49.2   & –      & 34.5    \\
    InstructBLIP (2023) 
      & Vicuna-13B  & 13.5  & 224  & 49.5   & –      & 33.4    \\
    IDEFICS-9B (2023) 
      & LLaMA-7B   & 9.0   & 224  & 38.4   & 50.9   & 35.5    \\
    IDEFICS-80B (2023) 
      & LLaMA-65B  & 80.0   & 224  & 45.2   & 60.0   & 36.0    \\
      \hline
      \textit{BcQLM (ours)}
      & \textit{Llama-3.2-1B} & \textbf{1.2}
      & \textit{224}  & \textbf{60.8}  & \textbf{71.0}  & \textbf{49.5} \\
      \hline
      \hline

    Qwen-VL (2023) 
      & Qwen-7B   & 9.6    & 448  & 59.3   & 78.8   & 35.2    \\
    Qwen-VL-Chat (2023) 
      & Qwen-7B   & 9.6   & 448  & 57.5   & 78.2   & 38.9    \\
    
    LLaVA-1.5 (2024) 
      & Vicuna-1.5-7B & 7.3 & 336  & 62.0   & 78.5   & 50.0    \\

     LLaVA-1.5 (2024) 
      & Vicuna-1.5-13B & 13.3 & 336  & \textbf{63.3} & 80.0   & 53.6    \\

    VILA-7B (2024) 
      & Llama-2-7B & 7.0 & 336  & 62.3   & 79.9   & 57.8 \\
    VILA-13B (2024) 
      & Llama-2-13B & 13.0 & 336  & \textbf{63.3}   & \textbf{80.8}   & \textbf{60.6} \\
      
      \hline
      \textit{BcQLM (ours)}
      & \textit{Llama-3.2-1B} & \textbf{1.2}
      & \textit{336}  & \underline{62.4}  & \textit{78.7}  & \textit{56.1} \\

    \midrule
    \bottomrule
  \end{tabular*}

  \caption{Comparison of accuracy (\%) across three VQA benchmarks: GQA, VQAv2, and VizWiz. 
Results for BLIP-2~\cite{blip2}, InstructBLIP~\cite{instructblip}, IDEFICS~\cite{IDEFICS}, 
LLaVA-1.5~\cite{llava1.5}, VILA~\cite{vila}, and Qwen-VL~\cite{qwen} are reported based on their official releases. 
Our proposed BcQLM achieves competitive performance across all datasets while using only 1.2B parameters. }
  \label{3} 
\end{table*}

To further improve the student model’s representations, we perform knowledge distillation from a pretrained CLIP model. Visual and textual embeddings $I_t$ and $T_t$ are extracted from the teacher model and projected into the student embedding space via learnable projection heads. The distillation loss is computed as the mean squared error (MSE) between the L2-normalised student and teacher embeddings:
\begin{equation}
\mathcal{L}_{\mathrm{distill}} = \frac{1}{\beta} \left( \mathrm{MSE}(I_s, I_t) + \mathrm{MSE}(T_s, T_t) \right),
\label{eq:distill}
\end{equation}
where $\beta$ is a normalisation factor.

The total training objective is defined as a weighted combination of the two losses:
\begin{equation}
\mathcal{L}_{\mathrm{total}} = \lambda_1 \cdot \mathcal{L}_{\mathrm{contrast}} + \lambda_2 \cdot \mathcal{L}_{\mathrm{distill}},
\label{eq:total}
\end{equation}
where hyperparameters, $\lambda_1$ and $\lambda_2$, balance the contributions of the contrastive and distillation losses.

\subsection{Multimodal Input Construction}



To enable fine-grained visual-language interaction, a Dynamic Gated Cross-Attention module is used. After extracting modality-specific representations from the feature encoders, visual and textual features undergo cross-modal integration to establish contextual grounding for answer generation. 

Specifically, the image encoder outputs a latent feature map \( F_V \in \mathbb{R}^{B \times C \times H \times W} \), which is spatially flattened into patch-level embeddings \( F_I \in \mathbb{R}^{B \times N \times C} \), where \( N = H \times W \). Concurrently, the text encoder provides token-wise embeddings \( F_T \in \mathbb{R}^{B \times T \times d} \), where \( d \) is the embedding dimension, as well as a global semantic representation \( \bar{F}_T \in \mathbb{R}^{B \times d} \) derived through average pooling:
\begin{align}
\bar{F}_T &= \frac{1}{T} \sum_{i=1}^T F_T^{(i)},
\end{align}

Visual patch tokens serve as queries, while textual embeddings provide keys and values for attention computation:
\begin{align}
Q &= F_I W^Q,\quad K = F_T W^K,\quad V = F_T W^V,
\end{align}
\begin{align}
\hat{F}_I &= \mathrm{Softmax}\left( \frac{Q K^\top}{\sqrt{d}} \right) V.
\end{align}

To inject global semantic control, the pooled textual vector \( \bar{F}_T \) is projected into the same latent space and broadcast across all spatial locations:
\begin{align}
\bar{F}_T' &= \text{Proj}(\bar{F}_T) \in \mathbb{R}^{B \times d},
\end{align}
\begin{align}
\gamma &= \sigma\left( \text{MLP} \left( [F_I ; \bar{F}_T'] \right) \right) \in [0,1]^{B \times N \times 1}.
\end{align}

The attention-enhanced visual representations \( \hat{F}_I \) are adaptively modulated by the gate \( \gamma \), then integrated with the original features through a gated residual connection:
\begin{align}
F_{\text{mod}} &= F_I + \gamma \odot \hat{F}_I,
\end{align}

This intermediate representation is subsequently refined via a feed-forward network and normalisation layer:
\begin{align}
F_{\text{fused}} &= \text{LayerNorm} \left( F_{\text{mod}} + \text{FFN}(F_{\text{mod}}) \right).
\end{align}

To align with the decoder’s token embedding space, the fused visual patches are mapped through a lightweight projection head, yielding pseudo textual tokens:
\begin{align}
F_{\text{pseudo}} = \text{Adapter}(F_{\text{fused}}) \in \mathbb{R}^{B \times N \times d}.
\end{align}


\subsection{LLM Decoder}

In this work, the LLaMA-3.2-1B model \cite{llama3} is adopted as the language decoder to generate final responses. As a compact yet capable member of the LLaMA series, the 1B variant is pre-trained exclusively on large-scale textual corpora, without exposure to any visual or multimodal data. Its small size enables efficient fine-tuning and inference, which makes it suitable for resource-constrained deployment. More importantly, the absence of any prior visual grounding makes it an ideal testbed to rigorously assess the effectiveness of our visual-text fusion module in aligning visual content for language generation.

Given the pseudo tokens from the image encoder, denoted as $F_{\text{pseudo}} \in \mathbb{R}^{B \times N \times d}$, and the instruction-response text sequence $X_{\text{text}}$, tokenised and embedded as $F_T = \text{Embed}(X_{\text{text}}) \in \mathbb{R}^{B \times T \times d}$, the input to the decoder is constructed by concatenation:
\begin{align}
F_{\text{llama}} = [F_{\text{pseudo}} ; F_T] \in \mathbb{R}^{B \times (N+T) \times d},
\end{align}

During training, label supervision is applied only to the response portion of the sequence, with instruction and visual tokens masked out from loss computation. This facilitates alignment of visual features with downstream language tokens via gradient propagation through the decoder:
\begin{align}
\mathcal{L}_{\text{gen}} = \text{CE}( \text{LLaMA}(F_{\text{llama}}), Y_{\text{response}} ).
\end{align}
where $\text{CE}(\cdot)$ denotes the cross-entropy loss computed over the generated response tokens $Y_{\text{response}}$.

\section{Experiments}

\subsection{Dataset}
To comprehensively evaluate the performance of our model, we conduct experiments on three widely used datasets: GQA, VQAv2, and VizWiz:
\begin{itemize}
  \item \textbf{GQA} \cite{GQA}:  
    113K real‐world images from Visual Genome paired with 22.7M programmatically generated questions. Each question is derived from a scene graph and converted into a natural‐language query covering comparison, spatial relations, logical inference and other multi‐step reasoning types.

  \item \textbf{VQAv2} \cite{vqav2}:  
    Built on MS-COCO, comprising approximately 205K images and over 1.1M image–question–answer triplets. Each question is paired with ten human‐provided answers and complementary image–question pairs are included to mitigate language biases.
  \item \textbf{VizWiz} \cite{vizwiz}:  
    Consists of about 31K images and corresponding free‐form questions captured by blind users with mobile phones. Each question has ten crowdsourced answers; the dataset is characterised by poor image quality, incomplete visual context and diverse real‐world query types.
\end{itemize}

\subsection{Settings}



We use a pretrained CLIP model as the teacher and BreezeCLIP as the student. Images are resized (\(224\times224\) or \(336\times336\)), normalised and converted to tensors. Text is tokenised and padded to 77 tokens. Visual and textual features are projected to a shared embedding space (512 or 768 dims). We use contrastive loss with \(\tau=0.07\), \(\alpha=\beta=0.5\), and train for 64 epochs using Adam (lr=\(1\times10^{-5}\)), StepLR (\texttt{step\_size}=10, \(\gamma\)=0.5), batch size 32, and gradient clipping (max norm=1.0).

\begin{figure}[t]
\centering
\begin{tikzpicture}

\begin{axis}[
    width=\columnwidth,
    height=7cm,
    xlabel={Epoch},
    ylabel={Cosine Similarity},
    grid=major,
    legend style={at={(0.4,-0.25)}, anchor=north, legend columns=2,legend cell align=left},
    smooth,
    thick,
    ymin=55, ymax=82,
    axis y line*=left,
    axis x line*=bottom
]

\addplot[color=blue, mark options={scale=0.5}] coordinates {
(1,57.44)(2,59.72)(3,62.84)(4,63.55)(5,64.68)(6,65.52)(7,65.56)(8,68.14)(9,69.21)(10,69.84)(11,70.13)(12,72.69)(13,72.90)(14,73.48)(15,74.40)(16,75.10)(17,76.00)(18,76.16)(19,76.18)(20,76.44)(21,76.50)(22,76.52)(23,76.56)(24,76.75)(25,76.84)(26,77.04)(27,77.30)(28,77.30)(29,77.34)(30,77.47)(31,77.52)(32,77.55)(33,77.64)(34,77.68)(35,77.97)(36,78.04)(37,78.16)(38,78.21)(39,78.24)(40,78.34)(41,78.41)(42,78.58)(43,78.67)(44,78.86)(45,79.06)(46,79.13)(47,79.47)(48,79.48)(49,79.51)(50,79.51)(51,79.52)(52,79.54)(53,79.55)(54,79.55)(55,79.56)(56,79.56)(57,79.59)(58,79.59)(59,79.60)(60,79.61)(61,79.62)(62,79.64)(63,79.66)(64,79.68)(65,79.72)(66,79.72)(67,79.75)(68,79.78)(69,79.83)(70,79.84)(71,79.89)(72,79.93)(73,79.94)(74,79.95)(75,79.96)(76,79.97)(77,80.02)(78,80.04)(79,80.06)(80,80.09)(81,80.24)(82,81.73)(83,80.09)(84,80.21)(85,80.25)(86,80.27)(87,80.22)(88,80.85)(89,80.17)(90,80.36)(91,80.54)(92,80.39)(93,80.18)(94,80.09)(95,80.22)(96,80.12)(97,80.29)(98,80.37)(99,81.13)(100,80.12)
};
\addlegendentry{BreezeCLIP Cosine}

\addplot[color=red, mark options={scale=0.5}] coordinates {
(1,71.09)(2,70.26)(3,70.31)(4,71.13)(5,70.27)(6,70.49)(7,70.34)(8,70.49)(9,70.50)(10,70.40)(11,70.25)(12,70.27)(13,70.37)(14,70.29)(15,70.39)(16,70.72)(17,71.00)(18,70.53)(19,70.36)(20,70.25)(21,71.20)(22,71.11)(23,70.49)(24,70.36)(25,70.27)(26,70.29)(27,71.08)(28,70.48)(29,70.51)(30,70.25)(31,71.17)(32,70.84)(33,70.22)(34,70.96)(35,70.23)(36,70.43)(37,70.29)(38,71.18)(39,71.13)(40,71.09)(41,70.25)(42,70.44)(43,70.40)(44,70.44)(45,71.17)(46,70.67)(47,71.15)(48,70.27)(49,71.22)(50,70.25)(51,70.37)(52,71.11)(53,70.23)(54,70.39)(55,70.29)(56,70.67)(57,70.92)(58,71.06)(59,70.72)(60,70.49)(61,70.33)(62,70.77)(63,70.34)(64,70.63)(65,71.20)(66,70.36)(67,70.31)(68,70.58)(69,70.51)(70,71.04)(71,70.43)(72,71.18)(73,70.58)(74,71.06)(75,70.27)(76,70.96)(77,71.15)(78,70.23)(79,70.63)(80,70.77)(81,70.84)(82,71.04)(83,70.23)(84,70.22)(85,70.53)(86,70.33)(87,70.21)(88,70.92)(89,71.22)(90,70.48)(91,70.36)(92,70.26)(93,71.00)(94,70.25)(95,70.44)(96,70.44)(97,71.08)(98,70.21)(99,70.27)(100,70.50)
};
\addlegendentry{Teacher CLIP Cosine}
\end{axis}

\end{tikzpicture}
\caption{Cosine similarity in BreezeCLIP distillation.}
\label{4} 
\end{figure}
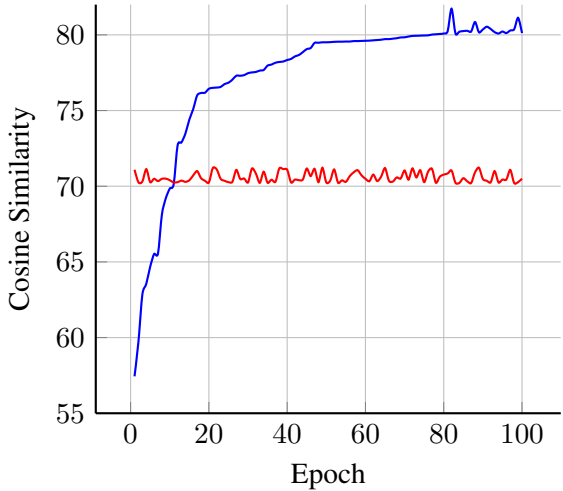

BreezeCLIP visual features are adapted via a VisualAdapter and fused with question embeddings through an 8-head Dynamic Gated Cross-Attention. The fused embeddings are fed into a trainable LLaMA model. Training uses AdamW (lr=\(1\times10^{-4}\)), StepLR (\texttt{step\_size}=5, \(\gamma\)=0.1), for 15 epochs, with batch size 32 and gradient clipping.

\subsection{Evaluation Metrics}

During the distillation and contrastive learning stage, we compute the InfoNCE loss on the validation set to measure the convergence of image–text embedding alignment.
Additionally, the mean cosine similarity of all positive image–text pairs and the positive–negative similarity gap are reported to evaluate the discrimination capability of the model. For the visual question answering task, we employ VQA accuracy as the sole evaluation metric, quantifying the proportion of generated answers that match the reference answers.

\subsection{Quantitative Results}
BreezeCLIP is jointly optimised with contrastive and distillation objectives under the supervision of a frozen CLIP ViT-B/32 teacher. During training, the cosine similarity between visual and textual embeddings increases steadily, ultimately exceeding 80\%, whereas the teacher model remains around 71\% (Figure~\ref{4}). This demonstrates that despite its compact size, BreezeCLIP achieves stronger alignment in the multimodal embedding space.

\begin{figure}[t]
  \centering
  \begin{subfigure}[t]{0.48\columnwidth}
    \centering
    \includegraphics[width=0.9\linewidth]{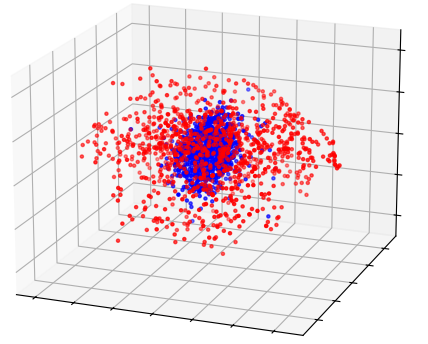}
    \caption{BreezeCLIP}
    \label{fig:breezeclip_tsne}
  \end{subfigure}
  \hfill
  \begin{subfigure}[t]{0.48\columnwidth}
    \centering
    \includegraphics[width=\linewidth]{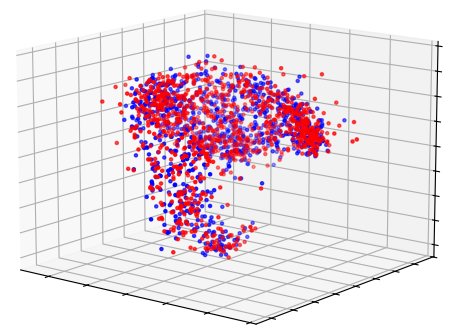}
    \caption{Teacher CLIP}
    \label{fig:teacherclip_tsne}
  \end{subfigure}
  
  \caption{3D projection of vision-language embeddings after dimensionality reduction. BreezeCLIP (a) exhibits clearer separation between positive and negative pairs compared to the teacher CLIP (b), which shows no separation between positive and negative pairs.}
  \label{5} 
\end{figure}

To better visualise the feature distribution, we project high-dimensional embeddings into 3D via PCA and show them in Figure~\ref{5}. The teacher model, although not trained on these datasets, still achieves about 71\% positive cosine similarity due to large-scale pretraining and the strong cross-modal representation learning, generalisation ability and robust alignment properties of CLIP. However, its embedding space mixes positive and negative pairs without clear separation. In contrast, BreezeCLIP clusters positive pairs tightly at the centre and pushes negative pairs outward. This demonstrates that our dual-loss training strategy significantly improves feature discrimination.

\begin{figure*}[h]
  \centering
  \includegraphics[width=\textwidth]{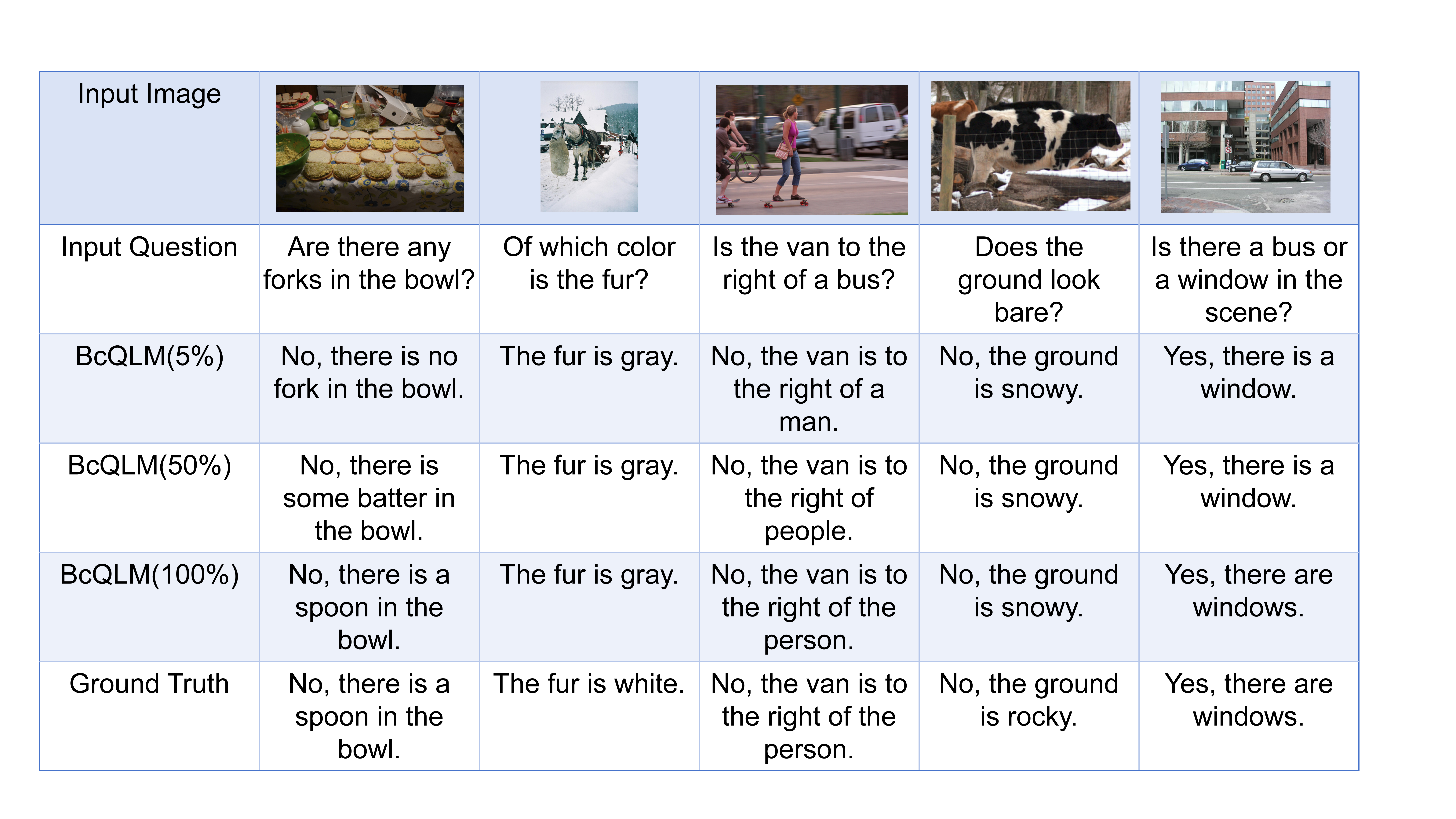}
  \caption{ Example BcQLM responses to GQA questions: “Is there a fork in the bowl?”, “What color is the fur?”,”Is the van to the right of a bus?“ “Is there a window in the scene?”, and “Does the ground look bare?” under three different LLaMA-3.2-1B unfreeze ratios (5\%, 50\%, 100\%). Despite unfreezing only 5\% of parameters, the model maintains correct semantic judgments; full (100\%) unfreezing further improves fine-grained detail recognition, such as correctly identifying a spoon in the bowl.}
  \label{6} 
\end{figure*}

The proposed framework is evaluated on three widely used VQA benchmarks (GQA, VQAv2 and VizWiz), using accuracy as the primary evaluation metric. As shown in Table~\ref{3}, BcQLM achieves an accuracy of 60.8\% on GQA, 71.0\% on VQAv2 and 49.5\% on VizWiz, outperforming most existing methods operating at a resolution of 224×224 and even surpassing several models using higher resolutions. When increasing the input resolution to 336×336, BcQLM improves its performance by achieving 62.4\% on GQA as the second best result among published models, 78.7\% on VQAv2 and 56.1\% on VizWiz as the third best result, bringing its overall performance in line with mainstream large scale vision language models. Notably, our model uses only 1.2B parameters, significantly smaller than existing models, and requires no external data or instruction tuning, demonstrating the effectiveness of our training and encoder design.

\subsection{Ablation Study}

Five configurations were compared on the validation set (as shown in Table~\ref{7}):
 CLIP + LLaMA-3.2-1B (58.8\%), BreezeCLIP + LLaMA-3.2-1B (59.2\%), Q-GCAM (60.8\%), Q-GCAM' with Token Balance (58.7\%) and Q-GCAM'' with Visual Query (55.7\%). Token Balance reweights visual features to match textual contributions; Visual Query adds a cross-modal attention layer using visual tokens as queries. These results show that Q-GCAM drives the performance gains, while its Token Balance and Visual Query variants need finer parameter tuning and integration.

\begin{table}[t]
\centering
\small
\begin{tabular}{c|c}
\toprule
\textbf{Model Configuration} & \textbf{Acc. (\%)} \\
\midrule
CLIP + LLaMA                     & 58.8 \\
BreezeCLIP + LLaMA              & 59.2 \\
BreezeCLIP + Q-GCAM + LLaMA      & \textbf{60.8} \\
BreezeCLIP + Q-GCAM' + LLaMA  & 58.7 \\
BreezeCLIP + Q-GCAM'' + LLaMA      & 55.7 \\
\bottomrule
\end{tabular}
\caption{Comparison of different model configurations on validation accuracy. All LLaMA means LLaMA-3.2-1B, Q-GCAM' indicates the Token Balance strategy, Q-GCAM'' indicates the Visual Query strategy.}
\label{7} 
\end{table}

In the preceding ablation study, we evaluate the impact of varying unfreezing ratios (5\%, 50\%, and 100\%) on BcQLM’s responses using the LLaMA-3.2-1B backbone. Across different settings, the model consistently produces semantically correct answers, even when only 5\% of the parameters are unfrozen. Fully unfreezing the model provides slight improvements on fine-grained details, such as object attributes and spatial descriptions. Importantly, every prediction can still be viewed as correct from certain interpretative perspectives, even though its expression may differ from the canonical ground truth. These findings demonstrate BcQLM’s robustness and effectiveness under different unfreezing strategies, with minimal reliance on extensive parameter updates. Example responses are shown in Figure~\ref{6} for qualitative reference.

\subsection{Efficiency Analysis}

We further evaluate inference efficiency on an NVIDIA RTX 4070 Ti using 20 test samples. 
BcQLM completes the batch in \textbf{2.54 seconds} (127.1 ms per sample), requiring 
\textbf{4989.5 MiB peak GPU memory} and \textbf{1.57$\times$10$^{11}$ FLOPs per sample}. 
In comparison, Qwen2.5-VL-3B \cite{Qwen2.5} takes 5.06 seconds (253.1 ms per sample), uses 7152.6 MiB memory, 
and costs 1.60$\times$10$^{11}$ FLOPs per sample, while Gemma3-4B \cite{Gemma3} takes 5.48 seconds 
(273.9 ms per sample), consumes 7432.8 MiB memory, and requires 1.85$\times$10$^{11}$ FLOPs per sample. 
These results show that BcQLM runs at \textbf{twice the speed} of Qwen2.5-VL-3B and Gemma3-4B, 
with around \textbf{30\% lower memory usage} and comparable or fewer FLOPs, 
highlighting its superior efficiency for edge-like deployment scenarios.

\section{Conclusion}

In this work, we present BcQLM, a lightweight multimodal language model, specifically designed for efficient and effective vision-language understanding under resource-constrained settings. Our approach is centred on BreezeCLIP, a compact vision-language encoder trained using a dual strategy of contrastive learning and knowledge distillation. It preserves cross-modal alignment while significantly reducing computational overhead. The model further incorporates Q-GCAM, a Q-Gated Cross-Modal Attention module to enable dynamic and context-aware fusion of visual and textual features for adaptive multimodal reasoning.

Our extensive experiments demonstrate that BcQLM achieves strong performance on three widely used VQA benchmarks — GQA (62.4\%), VQAv2 (78.7\%), and VizWiz (56.1\%) — using only 1.2 billion parameters, a fraction of the size of comparable models. These results validate the effectiveness of our lightweight design in balancing computational efficiency and high performance. Beyond its strong quantitative results, BcQLM offers practical advantages for real-world deployment. Its compact architecture enables efficient inference on modest hardware, which makes it suitable for applications in edge computing, assistive technologies and real-time multimodal systems. Our modular framework also allows for flexible adaptation to other multimodal tasks as a versatile solution for resource-limited environments.


\section*{Limitations}

While our work introduces an efficient and effective multimodal large language model under constrained computational settings, certain areas offer opportunities for further enhancement. First, our model relies on publicly available vision-language datasets, which may not fully capture the complexities and distributions of real-world multimodal scenarios. The limited diversity and quality of these datasets could constrain the model’s generalisation ability, particularly in domains requiring fine-grained or specialised reasoning. Second, the decoder (LLaMA-3.2-1B) employed in our system is pretrained solely on text and remains frozen during fusion training. Although this design choice enhances efficiency and controllability, it may also limit the model’s capacity to adaptively integrate visual semantics during generation. Finally, although our approach has been primarily validated on VQA benchmarks, future work should explore its extension to more dynamic modalities, including video, audio and real-time interactive settings.



\bibliography{custom}

\appendix
\newpage
\renewcommand{\thefigure}{A\arabic{figure}}
\setcounter{figure}{0}

\section{Additional Visualization}
\label{sec:appendix}

\begin{figure}[ht]
  \centering
  \includegraphics[width=\columnwidth]{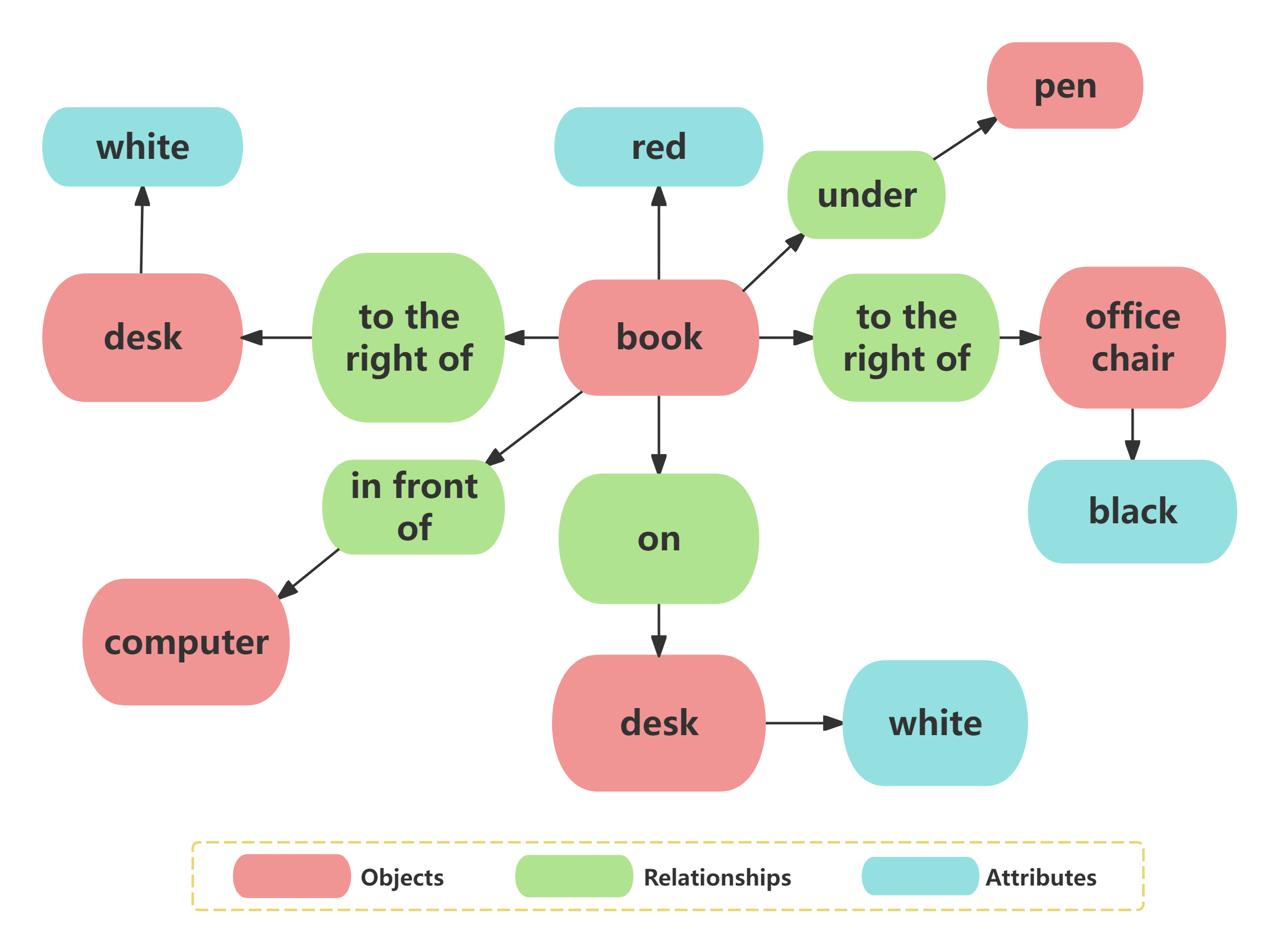}
  \caption{An example scene graph from the GQA dataset, illustrating spatial and attribute relationships among objects, such as \textit{a red book on the desk}.}
  \label{fig:appendix-sg}
\end{figure}

\end{document}